\documentclass[runningheads]{llncs}
\usepackage[T1]{fontenc}

\usepackage{graphicx,verbatim}

\usepackage{hyperref}
\usepackage{color}

\usepackage{amssymb}
\usepackage{bbm}
\usepackage[misc]{ifsym}
\usepackage{multirow}
\usepackage{pifont}
\usepackage[normalem]{ulem}
\useunder{\uline}{\ul}{}
\usepackage{xspace}

\newcommand{\ours}{NeuroLIP\xspace}

\begin{document}

\title{\ours: Interpretable and Fair Cross-Modal Alignment of fMRI and Phenotypic Text}
\titlerunning{\ours: Cross-Modal Alignment of fMRI and Phenotypic Text}

\author{
    Yanting Yang\inst{1,2}
    \and
    Xiaoxiao Li\inst{1,2}\textsuperscript{(\Letter)}
}
\authorrunning{Y. Yang and X. Li}
\institute{
    The University of British Columbia, Vancouver, BC V6T 1Z4, Canada \\
    \email{xiaoxiao.li@ece.ubc.ca}
    \and
    Vector Institute, Toronto, ON M5G 0C6, Canada
}

    
\maketitle              

\begin{abstract}
Integrating functional magnetic resonance imaging (fMRI) connectivity data with phenotypic textual descriptors (\emph{e.g.,} disease label, demographic data) holds significant potential to advance our understanding of neurological conditions. However, existing cross-modal alignment methods often lack interpretability and risk introducing biases by encoding sensitive attributes together with diagnostic-related features. In this work, we propose \ours, a novel cross-modal contrastive learning framework. We introduce text token-conditioned attention (\texttt{TTCA}) and cross-modal alignment via localized tokens (\texttt{CALT}) to the brain region-level embeddings with each disease-related phenotypic token. It improves interpretability via token-level attention maps, revealing brain region-disease associations. To mitigate bias, we propose a loss for sensitive attribute disentanglement that maximizes the attention distance between disease tokens and sensitive attribute tokens, reducing unintended correlations in downstream predictions. Additionally, we incorporate a negative gradient technique that reverses the sign of \texttt{CALT} loss on sensitive attributes, further discouraging the alignment of these features. Experiments on neuroimaging datasets (ABIDE and ADHD-200) demonstrate \ours's superiority in terms of fairness metrics while maintaining the overall best standard metric performance. Qualitative visualization of attention maps highlights neuroanatomical patterns aligned with diagnostic characteristics, validated by the neuroscientific literature. Our work advances the development of transparent and fair neuroimaging AI.


\end{abstract}

\section{Introduction}

Neuroimaging research increasingly utilizes phenotypic text, such as diagnostic labels and demographic descriptors, as semantic anchors to understand functional connectivity patterns~\cite{greene2022brain}. The integration of fMRI functional connectivity data with textual phenotypic descriptors offers significant potential to improve the prediction of neuropsychiatric disorders~\cite{gao2019impairments,lau2019resting}. Advances in cross-modal constrastive learning (CMCL), notably CLIP~\cite{CLIP} and SigLIP~\cite{SigLIP}, have successfully aligned visual and textual representations. These methods use constrastive learning strategies to learn joint embedding spaces between images and their corresponding text descriptions by training image and text encoders for image-level and text-level representations, and then maximize similarity between paired samples while minimizing similarity with negative pairs. Although these approaches have shown remarkable success in general computer vision tasks, their direct application to neuroimaging presents unique challenges, 
as they tend to obscure localized image-text relationships that are crucial for clinical interpretation, particularly in studying heterogeneous conditions like autism spectrum disorder (ASD), where associating brain regions with different text tokens is essential~\cite{isakoglou2023fine}. Moreover, CMCL in medical applications raises significant fairness concerns. Standard CMCL frameworks have been shown to inadvertently amplify demographic biases, such as encoding sex-related connectivity differences as proxies for psychiatric diagnoses~\cite{park-etal-2024-contrastive}. 
Therefore, the direct application of the existing CMCL methods to neuroimaging and phenotypic descriptors encounters two main challenges: (1) limited fine-grained interpretability of cross-modal associations (\emph{e.g.,} how text tokens are associated with brain regions) and (2) unintended demographic biases that conflate sensitive attributes with disease signatures~\cite{abbasi2020risk}.

Our work addresses these challenges by introducing a novel learning framework \ours{} that employs a \emph{\underline{t}ext \underline{t}oken-\underline{c}onditioned \underline{a}ttention} (\texttt{TTCA}) mechanism (Sec~\ref{sec:LTTC}) and a \textit{\underline{c}ross-model \underline{a}lignment via \underline{l}ocalized \underline{t}okens} (\texttt{CALT}) strategy (Sec~\ref{sec:sigmoid}) that learns alignment between each brain region with each text token. Unlike prior approaches that treat interpretability and fairness separately, we propose their joint optimization through our text token attention and contrastive learning pipeline (Sec~\ref{sec:Fairness}). 
Our innovative design achieves three key objectives: (1) developing an accurate neurodisorder classification model that utilizes both fMRI and phenotypic data; (2) mitigating bias by disentangling disease-related features from sensitive attributes using the \texttt{TTCA} module and a negative gradient loss approach in \texttt{CALT}; and (3) enhancing interpretability by localizing brain regions linked to disease prediction. 
We conduct comprehensive experiments on two independent neurodisorder datasets (ABIDE [ref] and ADHD-200 [ref]) to demonstrate our superior prediction performance and reduced bias among subgroups. Using attention map visualization and Neurosynth~\cite{Neurosynth} meta-analytic maps for psychiatric phenotypes, we also offer neurally plausible explanations that were absent in previous CMCL methods.

\section{Method}

\begin{figure}[b]
\centering
\includegraphics[width=\textwidth]{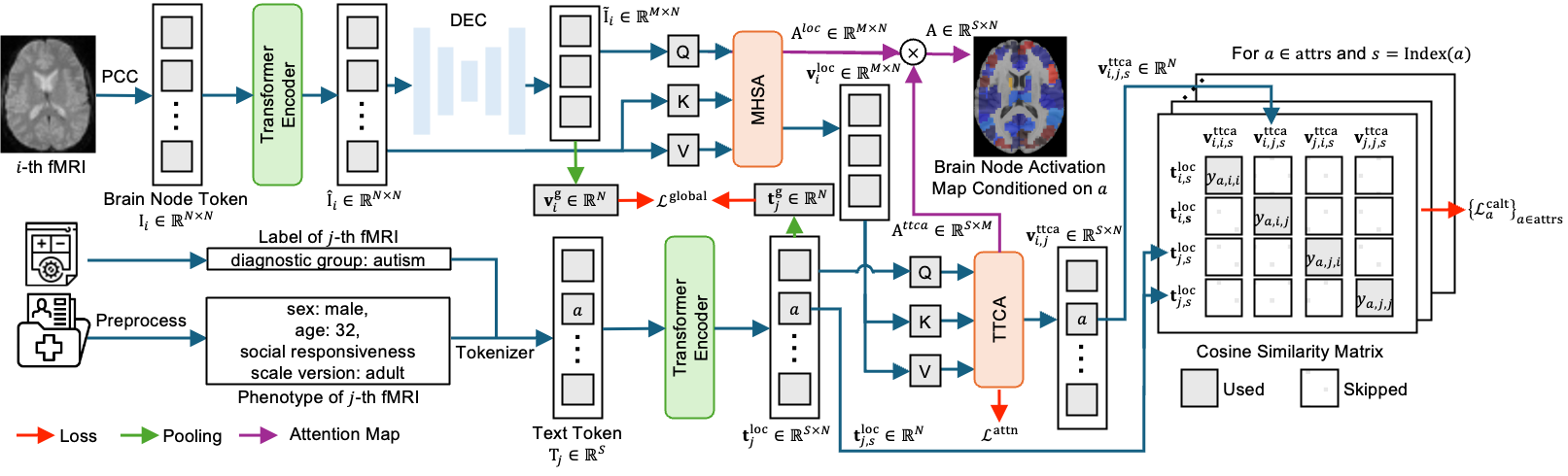}
\caption{Overview of \ours. Here, we show the pipeline applied on a batch of two ($i$-th and $j$-th) fMRI images and we present the interaction between the $i$-th image and the phenotypic text of $j$-th image.}
\label{fig:main}
\end{figure}

\subsection{Overview of \ours{} and Notations}

The overview of our proposed network is shown in Fig.~\ref{fig:main}. Given an atlas with $N$ ROIs, we obtain the average time series data from the voxels within each ROI in the $i$-th fMRI image and calculate Pearson's correlation coefficient (PCC) between time series of ROI pairs to measure connectivity, derived as $\mathbf{I}_i \in \mathbb{R}^{N \times N}$. The connectivity data $\mathbf{I}_i$ is input into the the vision model to obtain the brain node embeddings $\mathbf{\hat{I}}_i \in \mathbb{R}^{N \times N}$. To learn a better image representations, we use a multi-head self-attention (MHSA) layer, which takes brain node embeddings $\mathbf{\hat{I}}_i$ as key and value. For the query, we use a DEC module~\cite{DEC} to encode $\mathbf{\hat{I}}_i$ into $\mathbf{\tilde{I}}_i \in \mathbb{R}^{M \times N}$ and use $\mathbf{\tilde{I}}_i$ as input, which outputs the local image tokens $\mathbf{v}^\mathrm{loc}_i \in \mathbb{R}^{M \times N}$. The global image tokens $\mathbf{v}^\mathrm{g}_i \in \mathbb{R}^N$ is pooled over DEC output $\mathbf{\tilde{I}}_i$. For the text input of the $j$-th image $\mathbf{T}_j \in \mathbb{R}^{S}$, where $S$ is the text sequence length, the text data is input into the text model to obtain the local text tokens $\mathbf{t}^\mathrm{loc}_j \in \mathbb{R}^{S \times N}$. The global text token $\mathbf{t}^\mathrm{g}_j \in \mathbb{R}^N$ is pooled over local text tokens $\mathbf{t}^\mathrm{loc}_j$. Subsequently, \texttt{TTCA} is employed for dynamic image feature selections such that $\mathbf{v}^\mathrm{ttca}_{i,j} = f(\mathbf{t}^\mathrm{loc}_j, \mathbf{v}^\mathrm{loc}_i)$ (detailed in Sec.~\ref{sec:LTTC}). Finally, we apply the \texttt{CALT} loss $\mathcal{L}^\mathrm{calt}$, incorporated with the negative gradient technique, for image-text feature alignment (detailed in Sec.~\ref{sec:sigmoid}) and attention loss $\mathcal{L}^\mathrm{attn}$ for sensitive attributes debiasing (detailed in Sec.~\ref{sec:Fairness}).

\subsection{Text Token-Conditioned Attention}
\label{sec:LTTC}

\ours{} aims to integrate both fMRI data and phenotypic information into a neurodisorder classification framework while simultaneously interpreting how brain region activities are associated with disease labels. To achieve this, we introduce a \textbf{Text Token-Conditioned Attention} (\texttt{TTCA}) mechanism that explicitly aligns individual brain region activations with corresponding phenotypic descriptors and disease labels.  Most existing CMCL methods~\cite{CLIP,SigLIP} lack this level of fidelity, as they typically represent the input image as a single global embedding, thereby losing regional specificity. Instead of encoding the entire fMRI scan into a single vector, we aim to preserve the fidelity of brain regions by maintaining their distinct representations.  Inspired by the recent work FLAIR~\cite{FLAIR}, which demonstrated that contextualizing local image representations with global text embeddings can enhance interpretability, we extend this idea further. Specifically, we preserve local image tokens $\mathbf{v}^\mathrm{loc} \in \mathbb{R}^{M \times E}$, where $M$ represents the number of brain regions and $E$ is the embedding dimension, ensuring that each token corresponds to a distinct brain region. However, unlike FLAIR~\cite{FLAIR}, which aligns image features with a global text embedding, we retain token-wise text representations $\mathbf{t}^\mathrm{loc} \in \mathbb{R}^{S \times E}$, where $S$ is the number of text tokens. This modification enables the formation of localized image-text token pairs, where each brain region's activation is explicitly associated with a specific text token. With this enhancement, our \texttt{TTCA} allows us to extract image features that are dynamically conditioned on each individual text token, rather than a global representation.  
To effectively contextualize the image representation with local text tokens, we introduce an \texttt{TTCA} layer $f$, which produces text token-conditioned image representations $\mathbf{v}^\mathrm{ttca}$ from local image tokens $\mathbf{t}^\mathrm{loc}$. We define it as
\begin{equation}
    f(\mathbf{t}^\mathrm{loc}, \mathbf{v}^\mathrm{loc}) = \mathrm{softmax} \left( \frac{\mathbf{t}^\mathrm{loc} W_q (\mathbf{v}^\mathrm{loc} W_k)^\top}{\sqrt{d}} \right) \mathbf{v}^\mathrm{loc} W_v
\end{equation}
where $W_q$, $W_k$, $W_v$ are the query, key, and value projection matrices. In practice, we utilize a multi-head self-attention (MHSA) layer and append an empty token to $\mathbf{v}^\mathrm{loc}$ to allow $\mathbf{t}^\mathrm{loc}$ to attend to it when not semantically related to $\mathbf{v}^\mathrm{loc}$.

The attention weights produced by the \texttt{TTCA} layer offer valuable insights into the interactions between local image tokens and local text tokens, enabling the visualization of activation maps that highlight how different textual elements influence specific regions of an image. In practice, we apply another MHSA layer to condition local image tokens on DEC outputs, denoting attention weight matrices from \texttt{TTCA} and MHSA layer as $\mathbf{A}^\mathrm{loc} \in \mathbb{R}^{M \times N}$ and $\mathbf{A}^\mathrm{ttca} \in \mathbb{R}^{S \times M}$, respectively. Multiplying these gives the token-node attention matrix  $\mathbf{A} \in \mathbb{R}^{S \times N}$. Each element $A_{s,n}$ in $\mathbf{A}$ represents the attention weight between the $s$-th text and the $n$-th brain node, quantifying each brain region's relevance to a specific text descriptor. To visualize the activation map for a text token at index $s$, extract the corresponding attention weights across all brain regions:
\begin{equation}
    \mathbf{A}_s = \{ A_{s,1}, A_{s,2}, \dots, A_{s,N} \}
    \label{eq:activation_map}
\end{equation}
which can then be overlaid on the image to highlight regions that are most responsive to the text token $s$. This visualization facilitates the interpretation of cross-modal alignment by revealing which parts of the brain are activated by specific textual information. Moreover, by analyzing these activation maps across different text tokens, we can identify patterns of hypoconnectivity, where certain image regions exhibit consistently low attention weights in response to particular types of text tokens. Such patterns may indicate areas of the image that are less semantically aligned with the corresponding textual descriptions, potentially uncovering underlying structures or attributes that are not prominently captured by the model. Eq. (\ref{eq:activation_map}) demonstrates how localized attention quantifies interdependencies between image and text tokens. By leveraging these attention weights for visualization, we gain a deeper understanding of the model's interpretability, enabling the identification of meaningful cross-modal relationships and highlighting potential areas of improvement in the alignment process.

\subsection{Cross-Modal Alignment via Localized Tokens}
\label{sec:sigmoid}

We begin by considering the cosine similarity matrix between image and text embeddings, where positive pair is represented as $\langle \mathbf{v}^\mathrm{g}_i, \mathbf{t}^\mathrm{g}_i \rangle$ and the negative pair as $\langle \mathbf{v}^\mathrm{g}_i, \mathbf{t}^\mathrm{g}_j \rangle$, with $i \neq j$. The alignment process is guided by a sigmoid loss function~\cite{SigLIP}, denoted as $\mathcal{L}^\mathrm{global}$. Let $s$ denote the index of attribute $a$ within the text sequence, encompassing sensitive, non-sensitive attribute, and disease label. By employing \texttt{TTCA} (Sec~\ref{sec:LTTC}), we expand the scope of image-text pairs to image-text local token pairs. Therefore, the positive pair becomes $\langle \mathbf{v}^\mathrm{ttca}_{i,i,s}, \mathbf{t}^\mathrm{loc}_{i,s} \rangle$, while the negative pairs are $\langle \mathbf{v}^\mathrm{ttca}_{i,j,s}, \mathbf{t}^\mathrm{loc}_{i,s} \rangle$, $\langle \mathbf{v}^\mathrm{ttca}_{i,j,s}, \mathbf{t}^\mathrm{loc}_{j,s} \rangle$, and $\langle \mathbf{v}^\mathrm{ttca}_{i,j,s}, \mathbf{t}^\mathrm{loc}_{k,s} \rangle$, where $\mathbf{v}^\mathrm{ttca}_{i,j,s}$ is the image embedding from the $i$-th image conditioned on the $s$-th text token of image $j$, $\mathbf{t}^\mathrm{loc}_{i,s}$ is the $s$-th local text tokens of image $i$, and $i \neq j \land i \neq k$. For simplicity, we omit negative pairs related to different tokens. Previous study~\cite{FLAIR} shows that only the pair $\langle \mathbf{v}^\mathrm{ttca}_{i,j,s}, \mathbf{t}^\mathrm{loc}_{j,s} \rangle$ significantly contributes to effective image-text alignment, allowing us to disregard the other two types of negative pairs to reduce computational costs. The utilization of localized text tokens increases the likelihood that identical text tokens are shared across multiple images, particularly for categorical attributes. Consequently, for each attribute $a$, we construct the labels $\mathbf{y}$ as
\begin{equation}
    y_{a,i,j} = 2 \cdot \mathbbm{1}\left( G_{a,i} = G_{a,j} \right) - 1
\end{equation}
where $y_{a,i,j}$ is the label for the pair $\langle \mathbf{v}^\mathrm{ttca}_{i,j,s}, \mathbf{t}^\mathrm{loc}_{j,s} \rangle$, $\mathbbm{1}$ denotes the indicator function and $G_{a,i}$ represents the ground truth of attribute $a$ for image $i$. With positive / negative pairs and labels, we define our \textbf{Cross-Modal Alignment via Localized Tokens} (\texttt{CALT}) loss as
\begin{equation}
    \mathcal{L}^{\mathrm{calt}}_{a,i,j} = \frac{1}{1 + e^{y_{a,i,j} \left( -t \langle \mathbf{v}^{\mathrm{ttca}}_{i,j,s}, \mathbf{t}^\mathrm{loc}_{j,s} \rangle \right)}},
\end{equation}
where $s$ is the index of attribute $a$ and $t$ is a learnable weight. To incorporate the negative gradient technique, we can simply reverse the sign of this sigmoid loss, enforcing the unalignment of the sentitive attribute.

\subsection{Sensitive Attribute Disentanglement}
\label{sec:Fairness}

In the analysis of medical images and text, it is essential to distinguish between disease-related information and sensitive attributes, such as sex or age, to prevent the model from inadvertently learning or propagating biases associated with these attributes. To achieve this, we propose maximizing the distance between the attention weights related to disease tokens and those associated with sensitive demographic tokens. This encourages the model to treat these two types of information as distinct, thereby reducing bias. As detailed in Sec.~\ref{sec:LTTC}, the attention weight from the local text token-conditioned attention pooling layer signifies the interaction between local text tokens at position $s$ and local image tokens. We identify and isolate sensitive attributes by extracting the corresponding attention weights at their positions, denoted as $\mathbf{A}^\mathrm{ttca}_\mathrm{disease}$ and $\mathbf{A}^\mathrm{ttca}_\mathrm{sensitive}$. The attention loss can then be defined as
\begin{equation}
    \mathcal{L}^\mathrm{attn} = \|\mathbf{A}^\mathrm{ttca}_{\mathrm{disease}} - \mathbf{A}^\mathrm{ttca}_{\mathrm{sensitive}}\|_2
\end{equation}
where the L2 norm (Euclidean distance) is used to calculate this distance. The overall loss function $\mathcal{L}$ integrates the attention loss $\mathcal{L}^\mathrm{attn}$ designed to maximize the aforementioned distance, in addition to the \texttt{CALT} loss $\mathcal{L}^\mathrm{calt}$ that ensure image-text alignment, and a global loss $\mathcal{L}^\mathrm{global}$. The implementation involves optimizing this joint loss function during training:

\begin{equation}
    \mathcal{L} = \mathcal{L}^\mathrm{global} + \sum^\mathrm{attrs}_a \mathrm{sign}(a) \mathcal{L}^\mathrm{calt}_a - \beta \mathcal{L}^\mathrm{attn},
\end{equation}
where $\mathrm{sign}(a)=1$ for sensitive attributes and $1$ otherwise. $\beta$ is a hyperparameter.

\section{Experiments and Results}

\subsection{Experimental Setup}

\noindent{\textbf{Datasets:}} We used two publicly available resting state fMRI datasets: ABIDE~\cite{ABIDE} and ADHD-200~\cite{ADHD-200}. For ABIDE, we used data processed by the C-PAC pipeline~\cite{cpac2013} and that passed quality assessments, resulting in 403 individuals with ASD and 468 controls. For ADHD-200, we used data processed by the Athena pipeline~\cite{bellec2017neuro}, excluding images lacking phenotypic information, which produced 356 individuals with Attention-Deficit / Hyperactivity Disorder (ADHD) and 582 controls. Multiple individual scans resulted in 525 ADHD images and 842 control images. In both datasets, signals were bandpass filtered without global signal regression, and functional connectivity matrices were defined using the Craddock 200 atlas~\cite{CC200}. Phenotypic text was formatted as ``diagnostic group: \{dx\_group\}, sex: \{sex\}, age: \{age\}, social responsiveness scale version: \{srs\},'' where \{dx\_group\}, \{sex\}, and \{srs\} are categorical, and \{age\} is rounded to the nearest integer. The \{srs\} value, either child or adult, is determined by whether \{age\} is below or above 18. We select \{srs\} as the sensitive attribute for ABIDE dataset and \{sex\} for ADHD-200 dataset. All other attributes are viewd as normal attribute.

\noindent{\textbf{Metrics:}} For \textit{classification performance}, we use Area Under the Receiver Operating Characteristic Curve (AUC), Accuracy (ACC), Sensitivity (SEN), Specificity (SPC). For \textit{fairness}, we use Demographic Parity Difference (DPD)~\cite{DPD}, Difference in Equalized Odds (DEOdds)~\cite{DPD}, Equity-Scaled AUC~\cite{ES-AUC}. 

\noindent{\textbf{Training Setup:}} We utilize the AdamW~\cite{AdamW} optmizer with $1 \times 10^{-4}$ initial learning rate, $1 \times 10^{-4}$ weight decay, 32 batch size and 64 epoches. We use a cosine scheduler with a 3\% warmup ratio. We retain 20\% of the ABIDE data and the original ADHD-200 test release ($\approx$12.5\% of the data) as test set. We let $\beta = 0.001$ and implement a 5-fold cross-validation (CV) on the remaining training set for every experiment. All data split is implemented in a stratified manner based on the diagnostic group.

\subsection{Experimental Results}

\noindent\textbf{Baseline Comparison:} We first compare \ours with transformer-based fMRI classification model BrainNetTF~\cite{BRAINNETTF} supervised trained using disease label and CMCL baselines, including CLIP~\cite{CLIP}, SigLIP~\cite{SigLIP}, and FairCLIP~\cite{FairCLIP}, a fair learning method using Sinkhorn distance to align the visual-text feature similarities across demographic groups. As the mean value of 5-fold CV shown in Table~\ref{tab:comparison}, \ours consistently outperforms these baselines in all fairness measures while maintaining top classification performance. 
\begin{table}[t]
\scriptsize
\centering
\caption{Comparison with baseline methods. The best results are \textbf{bolded} and the second best results are {\ul underlined}.}
\label{tab:comparison}
\begin{tabular}{c|ccccccc}
\hline
\multirow{2}{*}{Method} & \multicolumn{7}{c}{ABIDE} \\ \cline{2-8} 
 & AUC ↑ & ACC ↑ & SEN ↑ & \multicolumn{1}{c|}{SPC ↑} & DPD ↓ & DEOdds ↓ & ES-AUC ↑ \\ \hline
BrainNetTF & 71.69 & 66.74 & 62.22 & \multicolumn{1}{c|}{{\ul 70.64}} & 13.12 & 21.40 & {\ul 69.04} \\
CLIP & 71.15 & 66.51 & 61.73 & \multicolumn{1}{c|}{{\ul 70.64}} & 15.60 & 22.92 & 67.77 \\
SigLIP & {\ul 71.88} & \textbf{67.77} & 64.20 & \multicolumn{1}{c|}{\textbf{70.85}} & {\ul 10.32} & 20.00 & 68.96 \\
FairCLIP & 71.09 & {\ul 67.09} & \textbf{67.90} & \multicolumn{1}{c|}{66.38} & 11.60 & {\ul 19.12} & 66.82 \\
\ours(ours) & \textbf{72.43} & 66.29 & {\ul 64.69} & \multicolumn{1}{c|}{67.66} & \textbf{7.92} & \textbf{16.70} & \textbf{70.00} \\ \hline
\multirow{2}{*}{Method} & \multicolumn{7}{c}{ADHD-200} \\ \cline{2-8} 
 & AUC ↑ & ACC ↑ & SEN ↑ & \multicolumn{1}{c|}{SPC ↑} & DPD ↓ & DEOdds ↓ & ES-AUC ↑ \\ \hline
BRAINNETTF & 55.48 & 58.60 & 37.40 & \multicolumn{1}{c|}{75.96} & 12.26 & 13.65 & 52.32 \\
CLIP & 59.12 & 59.18 & {\ul 42.08} & \multicolumn{1}{c|}{73.19} & 11.65 & 11.73 & 56.26 \\
SigLIP & {\ul 59.97} & \textbf{60.82} & \textbf{42.34} & \multicolumn{1}{c|}{75.96} & {\ul 9.88} & {\ul 10.49} & {\ul 57.63} \\
FairCLIP & 57.98 & {\ul 60.35} & 37.14 & \multicolumn{1}{c|}{\textbf{79.36}} & 11.53 & 10.85 & 55.76 \\
\ours(ours) & \textbf{62.74} & 60.12 & 38.96 & \multicolumn{1}{c|}{{\ul 77.45}} & \textbf{8.59} & \textbf{6.18} & \textbf{59.38} \\ \hline
\end{tabular}
\vspace{-2pt}
\end{table}

\noindent\textbf{Ablation Study:} To understand the impact of each component in \ours, we conducted a series of ablation studies by systematically removing key elements of our approach. As presented in Table~\ref{tab:ablation}, we evaluate the effect of the attention loss ($\mathcal{L}^\mathrm{attn}$) and the negative gradient (Neg. Grad.). The results indicate that both components contribute notably to the model's ability to mitigate bias. Without the attention loss, the fairness metrics, such as DPD and DEOdds, worsen, demonstrating its role in disentangling sensitive attributes from disease-related features. Similarly, the absence of the negative gradient technique results in higher bias, as indicated by increased fairness metrics, underscoring its importance in discouraging the alignment of sensitive attributes. The full implementation of \ours, combining both components, achieves the best balance between accuracy and fairness, confirming the synergistic effect of our proposed modifications in advancing fairness of neuroimaging analysis.

\begin{table}[t]
\scriptsize
\centering
\caption{The impact of each component of our proposed \ours. The best results are \textbf{bolded} and the second best results are {\ul underlined}.}
\label{tab:ablation}
\begin{tabular}{cc|ccccccc}
\hline
\multicolumn{2}{c|}{Component} & \multicolumn{7}{c}{ABIDE} \\ \hline
$\mathcal{L}^\mathrm{attn}$ & Neg. Grad. & AUC ↑ & ACC ↑ & SEN ↑ & \multicolumn{1}{c|}{SPC ↑} & DPD ↓ & DEOdds ↓ & ES-AUC ↑ \\ \hline
\ding{55} & \ding{55} & {\ul 72.32} & {\ul 67.66} & 62.47 & \multicolumn{1}{c|}{\textbf{72.13}} & 11.60 & 18.54 & 68.77 \\
\ding{51} & \ding{55} & 70.58 & 67.54 & {\ul 63.21} & \multicolumn{1}{c|}{{\ul 71.28}} & 10.48 & 18.32 & 67.59 \\
\ding{55} & \ding{51} & 72.10 & \textbf{68.11} & \textbf{64.69} & \multicolumn{1}{c|}{71.06} & {\ul 9.36} & {\ul 18.28} & {\ul 69.90} \\
\ding{51} & \ding{51} & \textbf{72.43} & 66.29 & \textbf{64.69} & \multicolumn{1}{c|}{67.66} & \textbf{7.92} & \textbf{16.70} & \textbf{70.00} \\ \hline
\multicolumn{2}{c|}{Component} & \multicolumn{7}{c}{ADHD-200} \\ \hline
$\mathcal{L}^\mathrm{attn}$ & Neg. Grad. & AUC ↑ & ACC ↑ & SEN ↑ & \multicolumn{1}{c|}{SPC ↑} & DPD ↓ & DEOdds ↓ & ES-AUC ↑ \\ \hline
\ding{55} & \ding{55} & 59.08 & 58.36 & 36.10 & \multicolumn{1}{c|}{76.60} & 9.26 & 12.34 & 54.93 \\
\ding{51} & \ding{55} & 58.89 & {\ul 58.60} & 35.84 & \multicolumn{1}{c|}{{\ul 77.23}} & 9.49 & 9.12 & 56.23 \\
\ding{55} & \ding{51} & {\ul 60.71} & {\ul 58.60} & \textbf{47.01} & \multicolumn{1}{c|}{68.09} & \textbf{7.85} & {\ul 8.22} & {\ul 58.79} \\
\ding{51} & \ding{51} & \textbf{62.74} & \textbf{60.12} & {\ul 38.96} & \multicolumn{1}{c|}{\textbf{77.45}} & {\ul 8.59} & \textbf{6.18} & \textbf{59.38} \\ \hline
\end{tabular}
\vspace{-10pt}
\end{table}

\begin{figure}[!htbp]
\centering
\includegraphics[width=\textwidth]{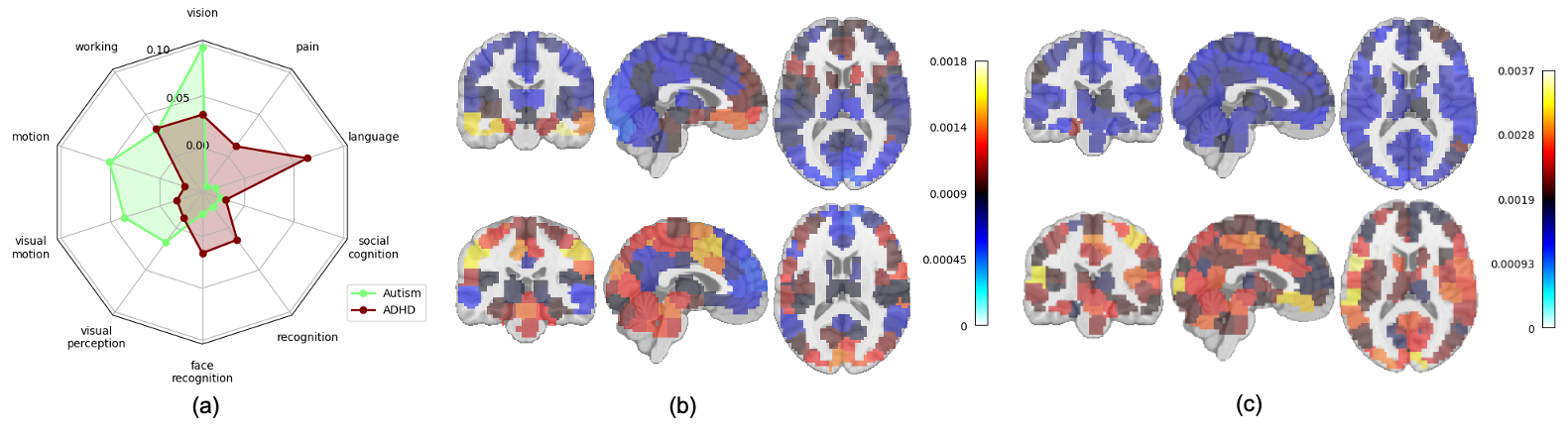}
\caption{(a) Meta-analysis. (b) Attention map conditioned on control (top) and ASD (bottom) in the ABIDE dataset. (c) Attention map conditioned on control (top) and ADHD (bottom) in the ADHD-200 dataset.}
\label{fig:vis}
\end{figure}

\noindent\textbf{Interpretation:} We visualize attention maps in Fig.~\ref{fig:vis} generated by our \texttt{TTCA} layer to illustrate how specific phenotypic descriptors modulate fMRI connectivity features. In both ABIDE and ADHD-200 datasets, these maps show distinct patterns when conditioned on diagnostic versus control tokens. In the ABIDE data set, autism-conditioned attention maps consistently highlight regions in the superior parietal lobes, which are key to spatial attention and sensory integration, often show altered engagement during visuospatial tasks~\cite{deramus2014enhanced}. Additionally, regions like the right middle and superior occipital gyri, along with the right lingual gyrus, support the processing of complex visual stimuli and object recognition, and are implicated in the local processing biases seen in autism~\cite{just2004cortical}. In the ADHD-200 datasets, the differential activation patterns in frontal operculum and middle frontal regions critical for inhibitory control and executive functions often show reduced activation during tasks requiring sustained attention~\cite{vetter2018anterior}. Bilateral insular cortices, which contribute to salience detection and the integration of emotional and cognitive information, are frequently reported as dysfunctional in ADHD, with evidence of both hypo- and hyperactivation~\cite{lopez2012reduced}. By comparing these maps against Neurosynth~\cite{Neurosynth} meta-analytic results, we observed that the activated brain regions of patients with Autism have a stronger correlation to vision~\cite{samson2012enhanced} and motion~\cite{van2019global}, while the activation map of patients with ADHD is more related to language~\cite{cohen2000interface} and recognition~\cite{romani2018face}, indicating a strong correspondence between our model’s highlighted regions and those frequently reported in the literature. This agreement reinforces the clinical plausibility of our findings.

\vspace{-8pt}

\vspace{-9pt}

\section{Conclusion}

In this work, we introduced \ours, a novel framework for cross-modal alignment that integrates fMRI connectivity data with phenotypic text. \ours significantly mitigates demographic bias while maintaining or improving classification performance. By leveraging \texttt{TTCA}, \ours generates interpretable attention maps that uncover neuroanatomically meaningful patterns. These maps, validated against meta-analytic data, provide clinicians and researchers with transparent insights into the underlying brain region-disease associations. Furthermore, our incorporation of a sensitive attribute disentanglement loss and a negative gradient technique effectively discourages the inadvertent alignment of demographic features with disease labels, addressing a critical challenge in neuroimaging analysis. Future work will focus on extending \ours to larger and more diverse datasets.

\clearpage
\newpage

\bibliographystyle{splncs04}
\bibliography{mybibliography}
\end{document}